\PassOptionsToPackage{svgnames}{xcolor}
\documentclass[journal]{vgtc}                     % final (journal style)
\onlineid{0}

\vgtccategory{Research}

\vgtcpapertype{please specify}

\title{SeeBel: \textbf{\underline{See}}ing is \textbf{\underline{Bel}}ieving}

\author{%
  \authororcid{Sourajit Saha$^*$}{0000-0003-1357-7813},
  \authororcid{Shubhashis Roy Dipta$^*$}{0000-0002-9176-1782},
}

\authorfooter{
  \item $^*$These two authors contributed equally to this work
  \item All authors are with the Department of Computer Science \& Electrical Engineering at the University of Maryland, Baltimore County.
}

\abstract{%
  Semantic Segmentation is a significant research field in Computer Vision. Despite being a widely studied subject area, many visualization tools do not exist that capture segmentation quality and dataset statistics such as a class imbalance in the same view. While the significance of discovering and introspecting the correlation between dataset statistics and AI model performance for dense prediction computer vision tasks such as semantic segmentation is well established in the computer vision literature, to the best of our knowledge, no visualization tools have been proposed to view and analyze the aforementioned tasks. Our project aims to bridge this gap by proposing three visualizations that enable users to compare dataset statistics and AI performance for segmenting all images, a single image in the dataset, explore the AI model's attention on image regions once trained and browse the quality of masks predicted by AI for any selected (by user) number of objects under the same tool. Our project tries to further increase the interpretability of the trained AI model for segmentation by visualizing its image attention weights. For visualization, we use Scatterplot and Heatmap to encode correlation and features, respectively. We further propose to conduct surveys on real users to study the efficacy of our visualization tool in computer vision and AI domain. The full system can be accessed at \url{https://github.com/dipta007/SeeBel}.
}

\teaser{
  \centering
  \includegraphics[width=\linewidth]{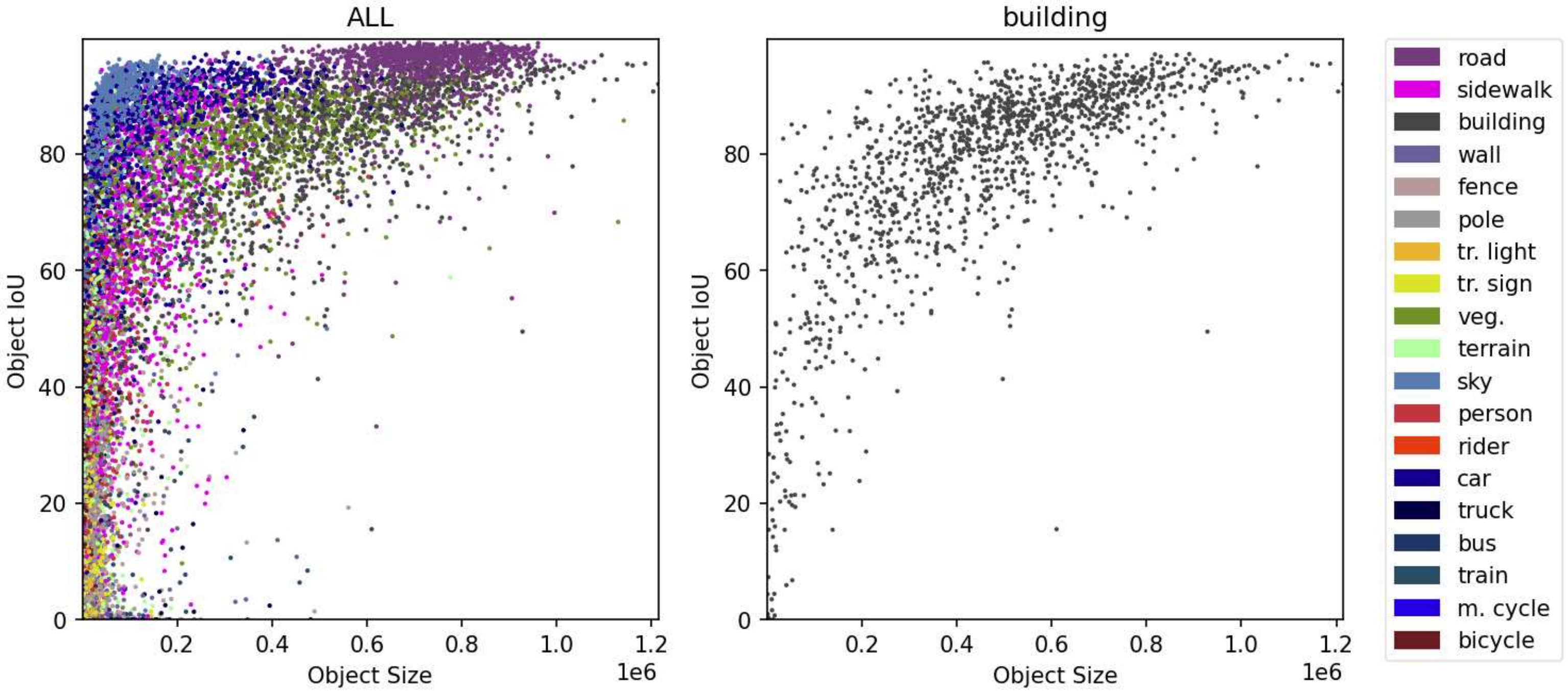}
  \caption{%
  	Correlation between object size and object performance for different object categories in cityscapes dataset \cite{cordts2016cityscapes} with an overview-detail view. In overview view, all object categories are shown, whereas in the detail view, specific object is shown.%
  }
  \label{fig:teaser}
}

\graphicspath{{figs/}{figures/}{pictures/}{images/}{./}} % where to search for the images
\usepackage{tabu}                      % only used for the table example
\usepackage{booktabs}                  % only used for the table example
\usepackage{lipsum}                    % used to generate placeholder text
\usepackage{mwe}                       % used to generate placeholder figures
\usepackage{algorithm}
\usepackage{algpseudocode}
\usepackage[colorinlistoftodos]{todonotes}
\usepackage{cite}
\usepackage{mathptmx}                  

\begin{document}

\maketitle

\section{Introduction}
In this project, we study how Data Visualization can aid machine learning systems users (researchers, engineers, practitioners) in analyzing the effects of class imbalance on semantic segmentation performance. Semantic segmentation is a high-level computer vision task where each pixel is assigned a class label delineating the object type (e.g., car, person, etc.) present in that pixel. Image datasets that are usually used to train machine learning models to perform semantic segmentation are imbalanced in terms of class distribution \cite{geiger2013vision, cordts2016cityscapes, lin2014microsoft, zhou2017scene, couprie2013indoor, mottaghi2014role, brostow2009semantic, pont20172017}. Moreover, class imbalance can negatively impact \cite{hossain2021dual} the performance of semantic segmentation systems. Therefore, investigating how class imbalance and the performance of semantic segmentation models are correlated is a significant research topic that requires investigation. The dataset that we use is Cityscapes \cite{cordts2016cityscapes} dataset. Our proposed visualization aims to visually assist machine learning systems users in inspecting the aforementioned correlation. Our proposed visualization consists of three key tasks: 

\subsection{User Discovers Correlation} \label{sec_1.1}
For this task, the user aims to discover correlation between class distribution and performance for semantic segmentation. We aim to design this visualization so that users can visually inspect prediction scores (by the trained model) and object size and compare these two attributes for all object classes in a given dataset. This task can potentially help users gain insights into how class distribution and AI performance relate to each other.

\subsection{User Locate/Explore/Browse Features}  \label{sec_1.2}
For this task, the user aims to locate, explore, or browse the model's attention weights across the entire image for each class, depending on the user's requirement. When segmenting an object, a trained AI model gives different amounts of attention to different portions of the image. With our visualization, the user can explore, browse, and locate through the image to find the amount of attention by AI at different image locations. Through this visualization, users can potentially gain insights into the AI model's behavior while predicting different objects.

\subsection{User Locate/Browse/Explore Distributions}  \label{sec_1.3}
For this task, the user's goal is to locate, browse or explore different object categories at different image regions as a trained AI model segments an image. This task will help the user visually/qualitatively inspect the segmentation quality of the model, and we propose an interaction for this task that allows users to bring all or any number of objects into view. Another objective of this task aims to help the user discover a quantitative correlation between object size and the confidence of the AI model to predict that object for a set of object categories that users can select. While the aim of the first task (described in section \ref{sec_1.1}) is to discover a correlation between object size and performance quantitatively for all images in the entire dataset, this task, however, does the same (both qualitatively and quantitatively) but for a single image.

\subsection{The problem potentially solved with our proposed visualization}
Our proposed visualization can potentially become a helping tool for researchers who train semantic segmentation models and individuals who are responsible for policy-making to deploy these systems in real-world applications. Semantic segmentation datasets are often imbalanced, and different techniques such as weighted loss function \cite{jozani2006estimation} and resampling data \cite{chawla2002smote} are usually used to mitigate these challenges. However, all of these techniques require an understanding of the dataset’s statistics. These statistics include the distribution of classes across a number of images (the number of images containing some object may be different than the number of images containing some other object), distribution of classes across the amount of pixel area (amount of pixel area containing some object may be different than the amount of pixel area containing some other object). With the knowledge of class distribution, the user can then design a loss function by assigning higher weights to the rare classes (low frequency, small object size) and lower weights to the abundant classes (high frequency, larger object size) before training the machine learning models. Sometimes, the aforementioned loss function's weights require an adaptive update. To understand in which direction an update is required (if at all), it is essential to know the classes' performance. And this knowledge can guide the user to change loss weights adaptively and learning rates or even to re-initiate training with a different approach if the performance for rare classes is still not improved, which can reduce both time, cost, and carbon footprint since the training of large machine learning models is expensive, time-consuming and power hungry. Our visualization, thus, can help the user depict these statistics, both before initiating the training and after training.

\subsection{Domain \& Typical Users}
The domain of our visualization is Machine Learning \& Computer Vision. Hence, some of the pronounced users of our visualization are Machine Learning (ML) and Computer Vision (CV) practitioners/researchers who might want to discover correlation between dataset statistics and performance. Another typical user is self-driving vehicle manufacturers who might want to discover how their trained AI model is doing for some crucial objects, i.e., roads, pedestrians, and traffic lights. The medical imaging community can be another set of typical users who might find it useful to discover the effect of dataset statistics on training, as medical data is often highly imbalanced.

\subsection{Our Contribution}
Our contributions are the following:
\begin{itemize}
    \item Our first visualization (detailed in section \ref{task_1}) helps the user discover correlation between class distributions and performance for each class.
    \item Our second visualization (detailed in section \ref{task_2}) allows the user to explore the attention weights of trained Semantic Segmentation models (AI) on images while predicting particular objects.
    \item Our last visualization (detailed in section \ref{task_3}) helps the user quantitatively discover correlations between object size and performance, both qualitatively and quantitatively for any single image and any number of object categories.
    \item Our visualizations can potentially help the users better understand their machine learning(AI) models and the effect of dataset statistics on training, contributing to enhancement in AI interpretability and building robust AI models.
    \item To the best of our knowledge, research efforts that are invested in building robust AI models do not typically use visualization tools to view dataset statistics and AI performance in a unified platform. While it is realized that dataset statistics and AI performance are closely related, a \textbf{one-stop} visualization tool to uncover that closeness is of sheer research significance.
\end{itemize}

\section{Related Works}
Some studies discovered the distribution of different classes in the cityscapes dataset in terms of the number of pixels occupied by each class \cite{cordts2016cityscapes, varma2019idd}, the proportion of labeled pixels for each class \cite{cordts2016cityscapes}, number of images for each class \cite{cordts2016cityscapes}, etc. These studies, however, are limited to analyzing (discovering) class distribution alone and do not compare any correlation between class distribution and performance. On the other hand, some studies - to depict the efficacy of their respective machine learning models - have reported \cite{zhang2021dcnas, takikawa2019gated} a detailed overview of performance for each class trained on various semantic segmentation datasets. Yet, they do not establish any connection between performance and class distribution. To the best of our knowledge, our visualization is the first to address a correlation between the aforementioned attributes.

Cost Curves \cite{drummond2006cost} is an essential visualization method to uncover machine learning model performance over a range of class distributions. The idiom for the cost curve is a line plot. The authors plot the probability cost of a wide range of sampling distributions over a dataset and their corresponding normalized expected cost. Moreover, to encode data, the authors have used horizontal and vertical spatial positions, connection lines, point marks, and shape channels for the user to find trends and correlations. However, generalizing the cost curve method to a semantic segmentation model is computationally (both memory and time) expensive since (1) cityscapes have 19 trainable classes, and the cost curve method uses only two classes to sample a wide range of distribution samples and (2) the complexity of a segmentation model is much higher than that of a classification model because. Furthermore, the cost curve is not easily interpretable from a visualization perspective. It takes domain-specific knowledge on the user’s end and added cognitive load to compare a correlation between a class distribution and model performance. Therefore, we only propose to report model performance and class distributions of the dataset using Scatter plots to discover their correlation in our project.

In Natural Language Processing (NLP), attention-based models have achieved state-of-the-art results in different domains of NLP tasks. But the interpretability of how they work remains a mystery. In this work \cite{vig2019multiscale}, the authors have used different idioms to increase the interpretability of attention-based models. The authors have provided three interactive idioms. In the attention-head view, authors have explored the self-attention weights for one or multiple heads for each input. They have used color hue to express the weight and line for word connections. Subsequently, in the model view, the authors have presented an overview-detail view with color hue as categorical layers and lines as word connections in a detailed view. And in the neuron view, authors have drawn a simple line connection idiom to show the individual neuron weights with respect to the query, key, and value. All of the views are interactive, with the option to choose layers and heads and zoom in on the individual head. Even though our proposal domain (computer vision) differs from this work's, the interpretability of their visualization has inspired us to study similar analogies for Computer Vision. 

Another study \cite{luque2021visualizing} was recently conducted, which proposed radial bar chart-based visualization of classification performance. They used the radial positions to represent model performance and the color and angular positions to represent classes. The authors further show how they draw different numbers of samples from a dataset and report the error rate for all of them after certain iterations. However, using such a design for our purpose will be cumbersome as we have more classes in cityscapes, and discovering trends from different figures (since it requires one visualization per iteration in this design mechanism) will increase the cognitive load on the user’s part. Due to the inefficiency of this proposed idiom, we have extensively used scatter plots in our visualization, which can show more points with less cognitive load.

A subsequent study \cite{colley2021effects} proposes a visualization that depicts the internal information processed by an autonomous vehicle so that the user, with a reduced cognitive load, can access that information and asses the vehicle's next move. It allows the users to perceive the vehicle’s detection capabilities, and the authors processed the semantic segmentation of road scene images for image classification based on pixels. They use color hue for categorical attributes to show the color distributions of the object visualization. Additionally, different shapes are used as identity channels to show dynamic and static objects. Furthermore, box plots and violin plots are used to show the distribution of data points. More specifically, violin plots are used to show the perceived ability to detect dynamic and static objects and the system's ranking. Whereas this visualization can successfully depict the information of autonomous vehicles, it has a limitation on the number of bins. Our dataset \cite{cordts2016cityscapes} has 19 classes, and a scatter plot is a more efficient choice than using a box plot.

In a later study \cite{raymaekers2022class}, the authors have proposed to learn insights into the training dataset from the corresponding classification results. Furthermore, this study aims to find how classification performance is related to class distribution by visually inspecting how far a predicted class is from the ground truth class in the class map space. The authors further show how they use scatter plots as their choice of idiom. The horizontal spatial position denotes class distance, and the vertical axis spatial position denotes the probability that the model assigns to alternative classes. \iffalse They have used color for categorical labels of the classification. \fi One of the major issues with this visualization is that it needs to be repeated for each class; therefore, using this mechanism to visualize the semantic segmentation performance for all the classes (19 of them) will be challenging, and discovering trends from multiple idioms will pose added cognitive load on the user’s part.

InstanceFlow \cite{puhringer2020instanceflow} is a sophisticated visualization tool that allows users to compare learning behavior between iterations on an instance level. They use a Sankey diagram with glyphs to depict a temporal analysis of the training process. They have used color to show the importance of the flow. With their proposed visualization, the authors offer a unified solution to discover trends between performance at different iterations at the expense of a complicated visualization and added cognitive load. We aim to allow the user to discover the exact correlation with a much simpler visualization (Scatter plot, images, and superimposed layer) with lesser strains on cognition.

\section{Implementation}
\subsection{Data}
The cityscapes \cite{cordts2016cityscapes} dataset has 5000 images with an identical resolution of 1024 × 2048 pixels. The dataset further contains masks for different objects in each image, and all of the images are taken while driving in different cities in Germany. Cityscapes contains 2975 training images, 500 validation images, and 1525 test images, contributing to a total of 5000 images. Moreover, the 1525 test images do not contain annotated masks. Those labels are not made public and are only used internally to score models submitted to the Cityscapes leaderboard \footnote{\url{https://www.cityscapes-dataset.com/benchmarks/}}. And therefore, we only use the training images for training our AI model and all relevant visualizations throughout the entirety of this project. The annotated masks provided in the dataset account for 30 classes. Although only 19 classes are used for evaluation, and therefore all of our experiments, including the pre-processing of the data and the corresponding visualizations, will contain 19 classes. 

\begin{figure}[t]
    \centering
    \includegraphics[width=\columnwidth]{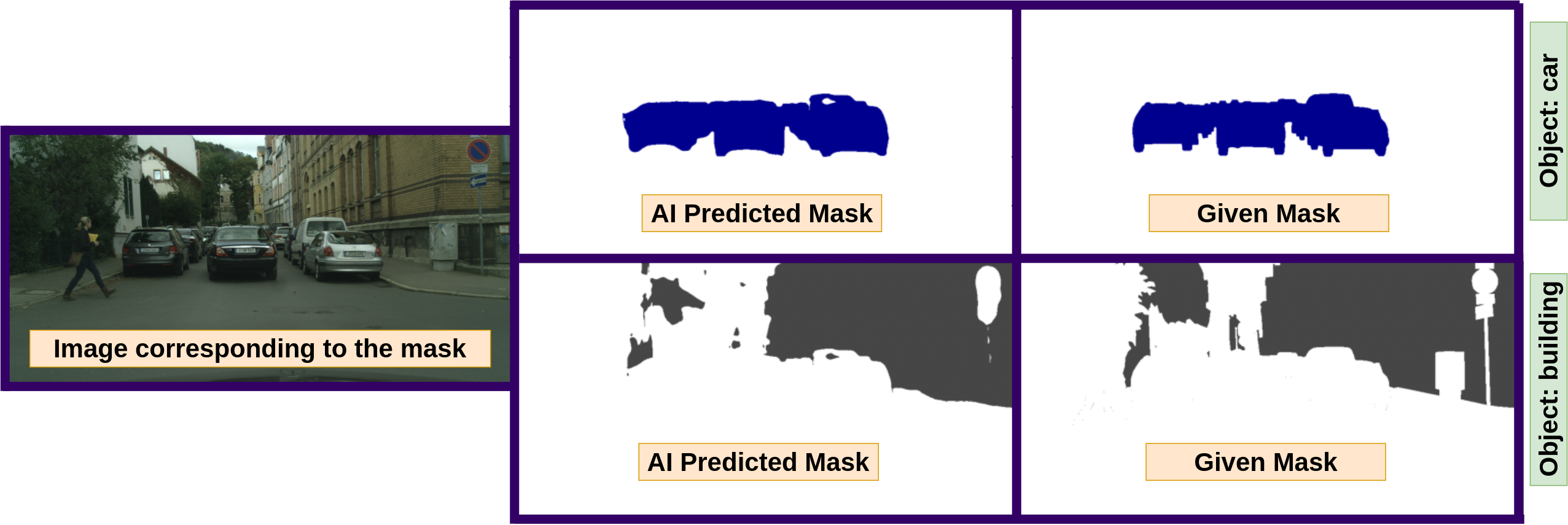}
    \caption{Input image along with it's given, and AI predicted masks for object category: car, building.}
    \label{fig:vis_1_trans}
\end{figure}

The semantics of this dataset are street scene images captured from vehicles. For every image in the dataset, there exists a set of corresponding masks that encompasses the objects present in that image and the position (pixel location) and category (i.e., car, bus, traffic light, etc.) of the object. Another way to think of this data is that all the pixels in an image contain an object of some kind (people, car, road, etc.). The dataset uses color hue (different color hues to represent different objects) for every pixel to describe what kind of object is present in that pixel.

\subsection{Data Transformation}   
The cityscapes dataset contains images (RGB images of scenes) and corresponding mask images. The dataset type for Cityscapes is geometry since images convey information about items with explicit spatial grid positions on a 2D surface. Therefore, the data type for cityscapes is items and spatial positions. One of the limitations of the Cityscapes dataset is the abundance of background classes meaning only a few objects of interest are labeled, leaving a lot of pixel space unlabeled (background class). Moreover, since we visualize the performance on the cityscapes dataset, one of the challenges in getting good performance on this dataset is the limited number of labeled training images \cite{lin2014microsoft, zhou2017scene}. Moreover, to denote the performance of AI models, we use the IoU (Intersection over Union) metric that signifies the overlap between the given mask and predicted mask for an object category in an image. The ideal range of IoU is $[0,1]$ \cite{Rezatofighi_2018_CVPR} however, for semantic segmentation IoU is traditionally \cite{wang2020deep, brostow2009semantic, couprie2013indoor, huang2019ccnet, wang2018understanding} expressed as a percentage and therefore the range becomes $[0,100]$ (multiplied by 100).

Now, there are multiple stages of transformation that we need to perform on the data. The first stage of transformation is task agnostic, and the second stage is task specific.

\subsection{Task Agnostic Data Transformation} \label{data_trans}
Firstly, since we need to get our data ready to be trained, we need to Resize the images from the original resolution (1024 x 2048) to 512 x 1024 pixels to match the computational capacity of our training hardware (NVIDIA RTX 3090 GPU). We use this image size for the remainder of this project in all of our experiments and analysis. Therefore, all the visualizations that use object size depend on the resized image rather than the original image. Training the AI model on the dataset is our project's significant data transformation process. Because once trained, we store the trained model to perform prediction later. Specifically, we have trained a Convolutional Neural Network based AI model called HRNet \cite{wang2020deep} on the cityscapes dataset. Furthermore, the hyper-parameters we have used for training are identical to this study \cite{wang2020deep}. Now, we perform these transformations because prediction masks are some of the attributes in our proposed visualizations, as described in section \ref{sec_1.1}, \ref{sec_1.2}, and \ref{sec_1.3}. 

\subsection{Task Specific Data Transformation for Task 1} \label{task1_trans}
Each image contains multiple objects (e.g., car, bus, road, etc.) and multiple masks. Consequently, a \textit{one-to-many} mapping exists between images and masks. Figure~\ref{fig:vis_1_trans} depicts two of many masks (mask of object category: car and building) for an image in the cityscapes dataset. To discover a correlation between object size and object IoU for different category classes, we transform the original dataset into a table using Algorithm \ref{alg:vis_1}. We iterate over all the masks ($m \in M$) in the original cityscapes dataset where $M$ is the set of all masks. Then, for each mask ($\forall m \in M$), we run the corresponding image, $i_m$, through an AI model ($AI$) to compute prediction mask $p$. Sequentially, we compute the IoU score ($IoU$) between each prediction $p$ and given mask $m$, object size of the given mask $m$ ($Size_m$), and object type of the given mask $m$ ($Object_m$) to construct each row of the table.

\begin{algorithm}[h]
\caption{Data transformation to construct a table from the original image dataset for the user to discover correlation among object size and object IoU for different object classes.}
\label{alg:vis_1}
\begin{algorithmic}
\State {$Table$ = []}
\For{$m \in M$}                                      \Comment{M is the set of all masks in the dataset}
    \State $p \gets AI(i_{m})$                   \Comment{Compute AI prediction}
    \State $ $                                       \Comment{for current mask $m$}
    \State $IoU \gets Score_{IoU}(p,m)$            \Comment{Compute IoU between AI prediction}
    \State $ $                                       \Comment{mask and current mask}
    \State $Size_m \gets Area_{pixel}(m)$            \Comment{Compute pixel area}
    \State $ $                                       \Comment{for current mask}
    \State $Object_m \gets ObjectType(m)$            \Comment{Extract object class of}
    \State $ $                                       \Comment{the current mask}
    \State $Table.add(Object_m$, $IoU_m$, $Size_m)$  \Comment{Add new mask to the table}
\EndFor
\State \Return {$Table$}
\end{algorithmic}
\end{algorithm}

\subsection{Task Specific Data Transformation for Task 2} \label{task2_trans}
To visualize the attention of the AI model for segmenting different objects, we utilize Grad-cam \cite{selvaraju2017grad}, an Explainable AI method. Grad-cam produces an image that shows the area of the input image to which the model is paying more attention than others, referred to as grad-cam weights. We transform the images (given in the dataset) to a superimposed view where the original images are superimposed with a grad-cam weight. The grad-cam algorithms must be called to retrieve AI attention weights to predict a particular object type. Now, the grad-cam algorithm requires a trained AI model ($AI$), image ($i$), and object category ($c$). We denote the grad-cam algorithm as $G$ and the grad-cam weight, $w = G(i, AI, c)$. Precisely, we repeat this process for all possible ($i, c$) pairs over the entire dataset $\forall i \in I$ and $\forall c \in C$. Here $I$ is the set of all images in the dataset, and $C$ is the set of all object categories in the dataset. Finally, to superimpose image $i$ and grad-cam weight $w$ together, we first display the original image $i$ with opacity $\alpha_1$. Then we display the grad-cam weight $w$ with opacity $\alpha_2$ to get the superimposed view $S$. This sequential process is described in Equation \ref{eq:trans_task_2}. Here, $\alpha_1 > \alpha_2$ so that the original image is visually accessible behind the grad-cam weight in the superimposed view.
\begin{equation}
\label{eq:trans_task_2}
S(i,w) = \{Display(i,\alpha_1) , Display(w, \alpha_2)\}
\end{equation}

\subsection{Task Specific Data Transformation for Task 3} \label{task3_trans}
We transform the images in the dataset into a table for this task with the help of the trained AI model ($AI$). We first run each image $i \in I$ in the dataset (where $I$ is the set of all images) through the trained AI model to get the predicted mask $p = AI(i)$. Then, we extract the given mask $m$ corresponding to image $i$. From both $m$ and $p$ we compute the number of pixels occupied by each object present in the corresponding masks. From the pixel count for each object category, we normalize these count values in the $[0,100]$ range to express these pixel counts in pixel occupancy percentage. Sequentially, we construct a table where each row is the current object's category in the image, the object's pixel occupancy in the given mask, and pixel occupancy in the predicted mask.

\subsection{Programming Language \& Visualization Tool}
We use Python3.7 \cite{python3} as our base programming language. We use Python Image Library (PIL) \cite{pil} to manipulate and display images in our visualization. For the visualization itself, we use matplotlib \cite{matplotlib}, GRAD-Cam \cite{gradcam}, and Jupyter Widgets \cite{ipywidgetsJupyterWidgets} for the interaction in our visualizations. Also, we used PyTorch \cite{pytorch} to train the machine learning model to predict the labels and mask. Also, for data manipulation, we used pandas \cite{pandas}, chardet \cite{pypiChardet}, and einops \cite{einopsEinops}.

\subsection{Software Architecture}
We use conda \cite{anaconda} and pip \cite{pip} package manager to install all of our dependencies for the code environment. We use JupyterLab \cite{jupyter} to run, host, and interact with the plotting. In our early alpha and beta release experiments, we tried other tools, i.e., seaborn \cite{seaborn}, tableau \cite{tableau}, but we switched to JupyterLab because we required more control over manipulating visualization, data transformation, and computational efficiency.

\section{Results}
\subsection{Task 1: Discovering Correlation Between Object Size and Object IoU for Different Category Classes} \label{task_1}

\subsubsection{Proposed Artifact}
\begin{figure}[t]
    \centering
    \includegraphics[width=\columnwidth]{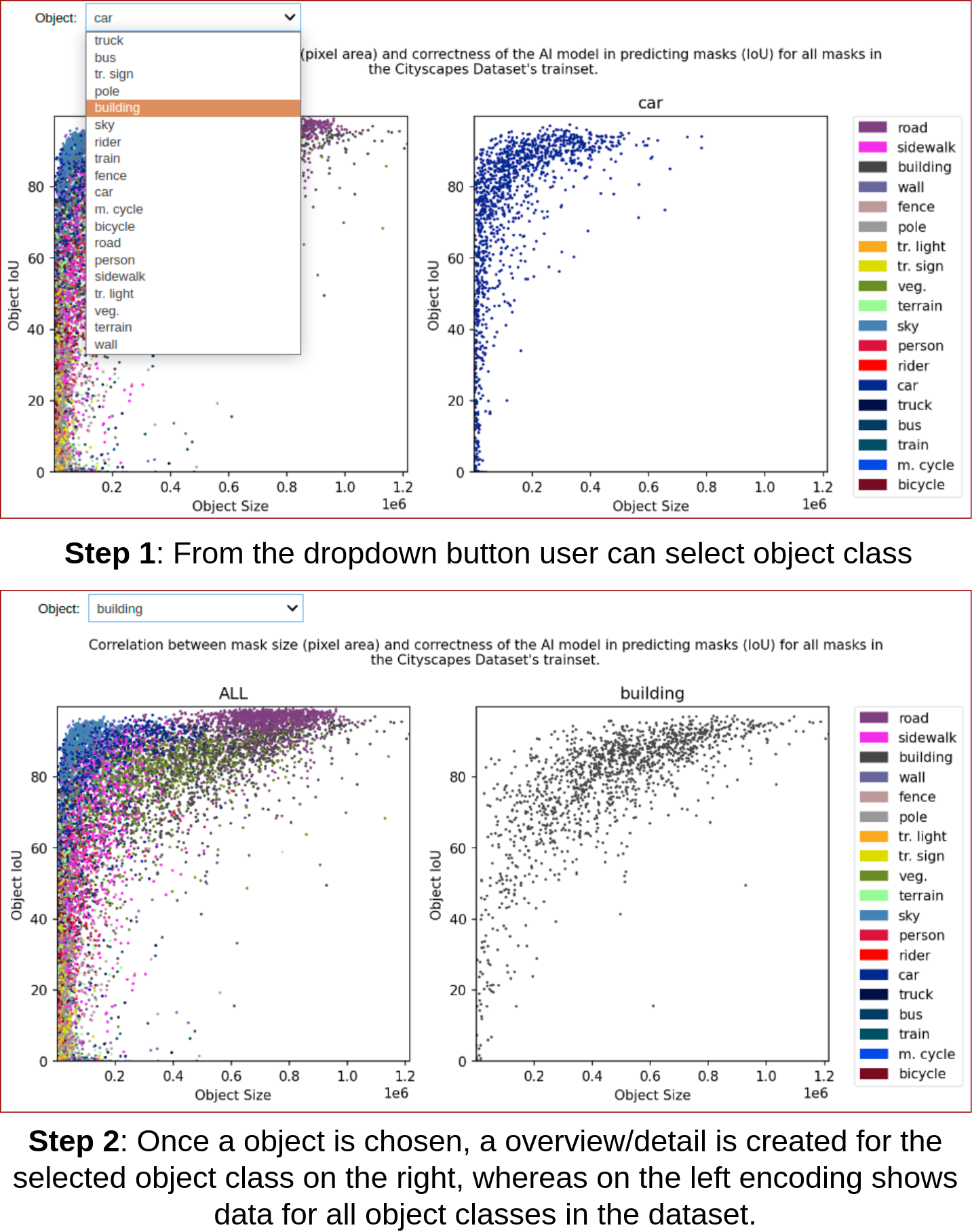}
    \caption{Snippet of our proposed tool for Task (section \ref{task_1}) from jupyter notebook. Top: a screen capture of how users can interact with the tool. The users can select an object class from the dropdown menu. At the initial condition when the user is yet to interact with the tool, on the left is displayed scatterplot for all objects (\textit{overview}), and on the right is displayed the scatterplot for car object (\textit{detail}). Down: screen capture of when the user selects building; on the left is displayed scatterplot for all objects (\textit{overview}), and on the right is displayed the scatterplot for building object (\textit{detail}).}
    \label{fig:vis_1}
\end{figure}

While using our proposed visualization tool (Figure~\ref{fig:vis_1}), the users' goal is to discover a correlation between object size and IoU for different category classes. Our proposed artifact uses two scatter plots with an overview/detail view, shown in Figure~\ref{fig:vis_1}. While our tool shows a scatter plot of object size and IoU for all object classes on the left (overview), the user, with the help of our dropdown menu, can select a particular object class and the scatter plot for the selected object category is then displayed on the right (detail). Figure~\ref{fig:vis_1} shows a snippet of how users can interactively use our tool for the aforementioned task.

\subsubsection{Data Abstraction}
The dataset type for this task is a 2D table with items and attributes as data types. In the aforementioned table, each item (row) is an object mask in the dataset. And there are three attributes (columns): (1) categorical attribute: object type (e.g., road, person), (2) ordinal attribute: object IoU (e.g., 55 \%, 92 \%), and (2) ordinal attribute: object size (number of pixels occupied). We use a horizontal position channel to encode object size, a vertical position channel to encode object IoU, and a color-hue channel to encode object category (class). Finally, we use a point mark for this idiom.

\subsubsection{Task Abstraction}
The user task is to \textbf{discover} a \textbf{correlation} between object size and object IoU for different category classes. These correlations can be positive, negative, reversing (first goes up and then down and vice-versa), or weak correlations.

\subsubsection{Encoding and Interaction Idiom}
For visual encoding, we use a scatter plot. Furthermore, we use overview/detail interaction with the same encoding but with a subset of the data. Finally, the dropdown menu we use to choose an object category is another form of interaction that falls under the select category. In the initial overview, our tool shows an overview scatter plot of object size and IoU for all object categories and a detail-view scatter plot of object size and IoU for the car object category (Figure~\ref{fig:vis_1}).

\subsubsection{Critical Analysis}
In \textit{overview} mode, our proposed tool (Figure~\ref{fig:vis_1}) is supposed to give a sense of correlation between object size and IoU across the whole dataset for all object categories to the user. However, it is difficult to track objects in \textit{overview} mode simply because there are too many masks in the dataset and, therefore, too many data points and colors in the scatter plot, as shown in Figure~\ref{fig:vis_1}. However, in \textit{detail} mode, it is much easier for the user to discover correlation because there is only a single object in the view, and the user is not shown data points for other object categories. Also, showing a scatter plot for a single object eliminates the need for color to denote object category, reducing a channel.

\begin{figure}[t]
    \centering
    \includegraphics[width=\columnwidth]{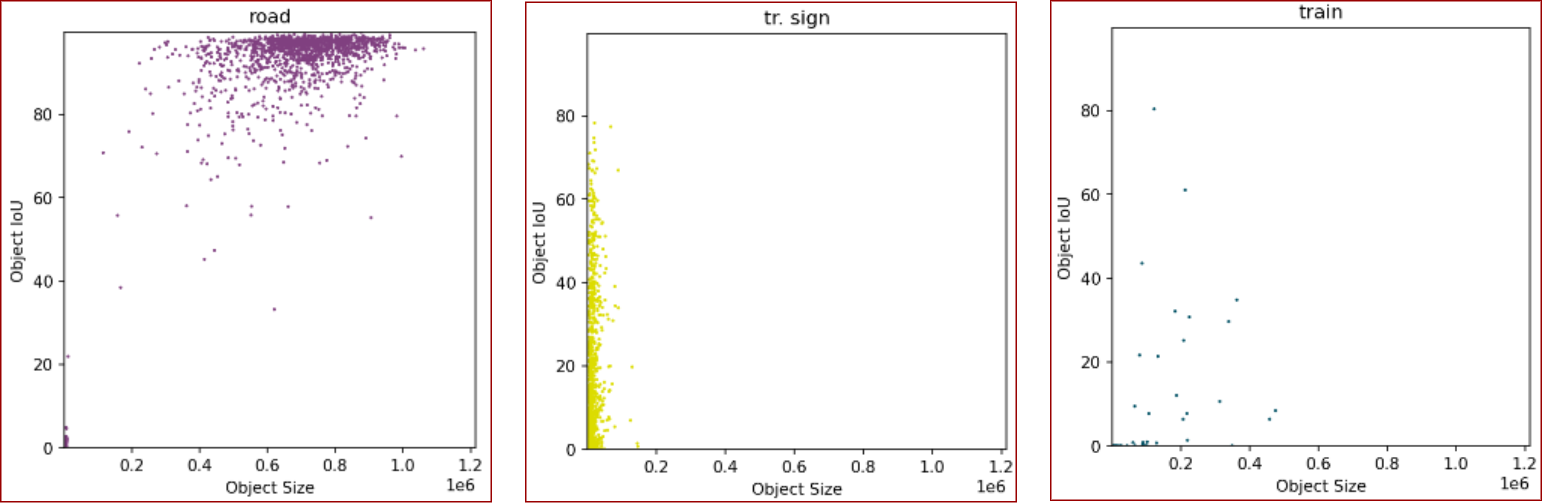}
    \caption{Cropped screenshot of our proposed tool for Task (section \ref{task_2}) from Jupyter notebook. Left: \textit{detail} view of the correlation between object size and IoU for road objects, Middle: \textit{detail} view of the correlation between object size and IoU for traffic sign objects, Right: \textit{detail} view of the correlation between object size and IoU for train objects.}
    \label{fig:vis_1_ana}
\end{figure}

Figure~\ref{fig:vis_1}, \ref{fig:vis_1_ana} further suggests that correlation can be observed using our proposed tool. For example, it can be observed (Figure~\ref{fig:vis_1_ana}) that for certain objects (e.g., road, train), there is no obvious correlation followed by all the masks. While for other objects (e.g., buildings), there is a positive correlation (Figure~\ref{fig:vis_1}). 

Interestingly, along with the goal of discovering correlation, it is further observed in Figure~\ref{fig:vis_1_ana} that from the density of the scatter plot, the user can also identify data distribution (e.g., there are fewer instances of train objects in the cityscapes dataset in comparison to that of traffic sign). Furthermore, our proposed visualization tool (Figure~\ref{fig:vis_1}, \ref{fig:vis_1_ana}) also helps the user identify other interesting distributions, such as most of the car objects sizes are below an upper limit of 0.6 Million, whereas most of the road object is sized within 0.4 to 1 Million and certain objects such as traffic signs are very tiny in size (within 0.05 Million).

On the other hand, one limitation of our design (Figure~\ref{fig:vis_1}) is the scalability issues that can arise with increased data because scatter plots are usually best suited for hundreds of data points. Another design limitation is the separability of objects with different color hues in \textit{overview} mode. For example, bus and train are not easily differentiable (Figure~\ref{fig:vis_1}) with the assigned colors, and as the number of object categories grows, the problem worsens.

Color hue denotes different object categories proposed in the cityscapes dataset  \cite{cordts2016cityscapes} are traditionally followed in computer vision literature \cite{wang2018understanding, huang2019ccnet, breitenstein2022amodal}. Since our target users are vision researchers, we have decided to honor that choice using the same color map used originally in the cityscapes dataset. We hope this decision will partly attract more target users to use our tool. Data transformation performed for this task (described in section \ref{task1_trans}) improves the match between data semantics and the task because discovering correlations between attributes (Object size and IoU) would not have been possible without the transformation. Additionally, we have used a horizontal position channel to denote object size, a vertical position channel to denote object IoU, and a color hue channel to denote object category. Removing any of these channels will disable us from showing all of these three attributes (object type, size, and IoU).

\subsection{Task 2: Locating, Browsing and Exploring Features} \label{task_2}

\subsubsection{Proposed Artifact}
\begin{figure*}[t]
    \centering
    \includegraphics[width=\textwidth]{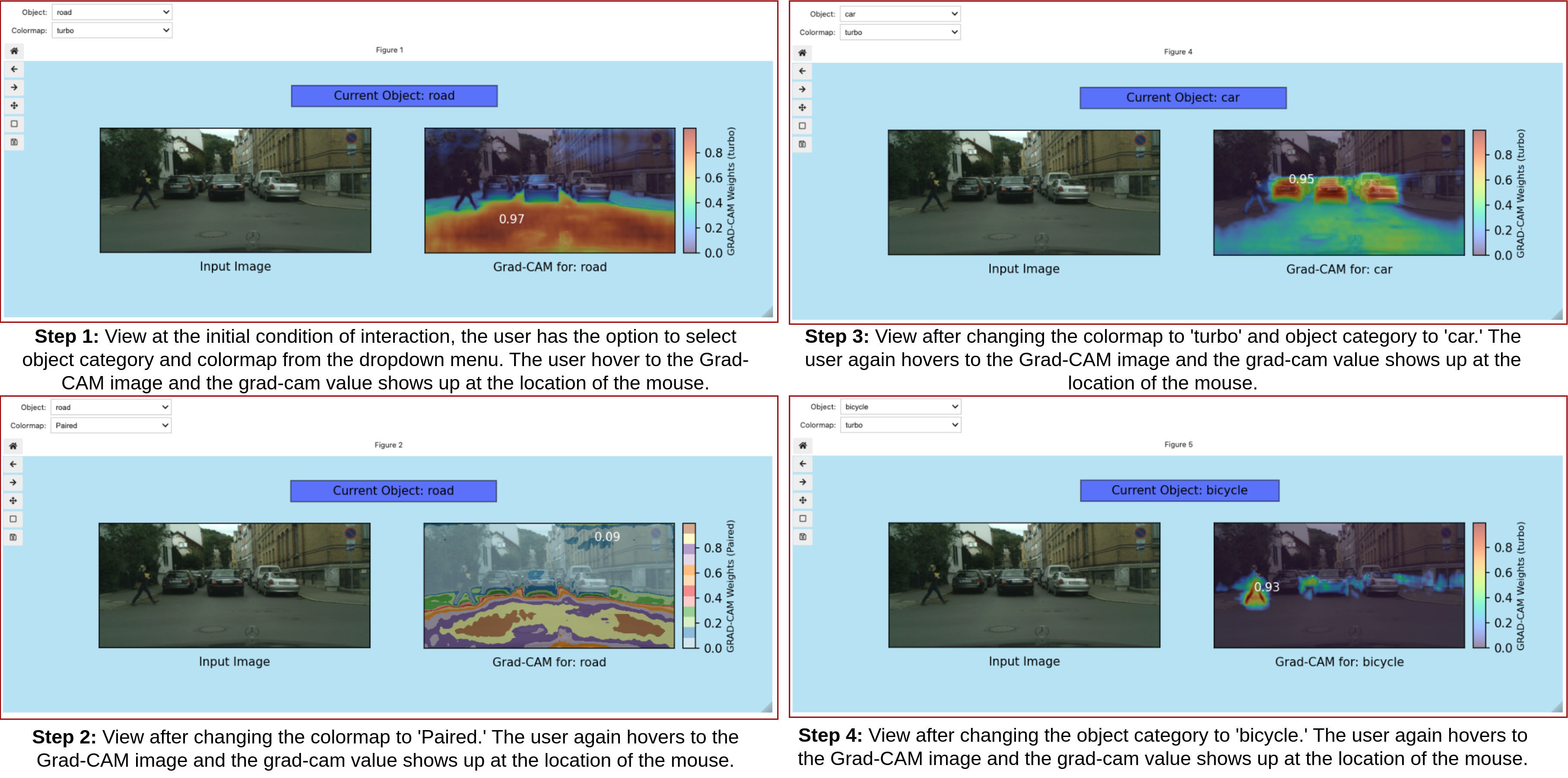}
    \caption{Screenshot of our proposed tool for Task (section \ref{task_2}) from Jupyter notebook. Top Left: A screen capture of the view where the user hovers over the GRAD-CAM image without interacting with colormap and object choices. Bottom Left: A screen capture of the view where the user only hovers over the GRAD-CAM image and changes the colormap. Top Right: A screen capture of the view where the user hovers over the GRAD-CAM image and changes both the colormap and object category. Bottom Right:  A screen capture of the view where the user hovers over the GRAD-CAM image and changes the object category only.}
    \label{fig:vis_2}
\end{figure*}

The goal of users with our proposed visualization tool (Figure~\ref{fig:vis_2}) for this task is to locate, browse, and explore grad-cam features. Our proposed artifact uses one image and one superimposed layer in a multiform view. As with our current design, the user cannot choose a different image for this tool; therefore, this tool runs with a static image chosen in the background. With the help of the dropdown menu, the user can select object category and colormap. The static RGB image is displayed on the left, and GRAD-Cam weights (superimposed with the static RGB image) are displayed on the right side of the multiform view. Moreover, the user can interact with the superimposed GRAD-CAM layer by hovering with the mouse, and the corresponding GRAD-CAM weight is annotated at the mouse location. Figure~\ref{fig:vis_2} shows an overview of our proposed tool for this task. GRAD-CAM \cite{selvaraju2017grad} weights signify the amount of importance given by the AI model to the image while predicting a certain object category.

\subsubsection{Data Abstraction}
The dataset type for task 2 is Spatial Field, and the data type is grid, position, and attribute. In the grid, each location denotes pixel coordinates while the attributes are pixel values. In the case of RGB images, the data can be viewed as 3 spatial fields where each spatial field's grid positions contain quantitative attributes such as R/G/B color values (different color channels at different spatial fields). There are several channels: (1) horizontal spatial position, (2) vertical spatial position, (3) color hue (when categorical color maps are used with color bins for the grad-cam superimposed layer), and (4) color saturation (when sequential color maps are used for the grad-cam superimposed layer or for showing images). Finally, we use the area as the mark for this idiom.

\subsubsection{Task Abstraction}
When the user hovers to a particular grad-cam superimposed layer position, their goal is to \textbf{browse} \textbf{features} (grad-cam weight). When the user looks at the color bar and wants to find a particular color in the grad-cam superimposed layer, the user wants to \textbf{locate} \textbf{feature} (either grad-cam weight or color). And finally, another use-case is when the user hovers through the grad-cam superimposed layer to \textbf{explore} \textbf{features} (grad-cam weight) to get a sense of the AI models' attention on different parts of the image.

\subsubsection{Encoding and Interaction Idiom}
We use a multiform view encoding with multiple interaction choices. As shown in Figure~\ref{fig:vis_2}, on the left of the multi-form view is an image, and on the right is a superimposed layer. One interaction choice is select (mouse hover) over the superimposed layer. Another form of interaction we use is highlighting; at the hovered mouse position of the superimposed layer, we display grad-cam weight at that position. Finally, the dropdown menu we use to choose object categories and colormaps is another form of interaction (select). In the initial overview, our tool shows an image, and it's a superimposed version with grad-cam weights for the road object category in the turbo color map.

\subsubsection{Critical Analysis}
Firstly, when the user hovers in the superimposed layer of our multiform-view tool, the corresponding grad-cam weight value is displayed (Figure~\ref{fig:vis_2}). This design is effective because the user - without needing additional interaction with or prompt on the tool - can browse grad-cam attentions at different image regions, and the view (superimposing image with the grad-cam weight) updates when the user hovers to a new location are relatively seamless. Moreover, the user can easily select object categories and colormaps from the dropdown menu to load into the view (Figure~\ref{fig:vis_2}). 

\begin{figure}[t]
    \centering
    \includegraphics[width=\columnwidth]{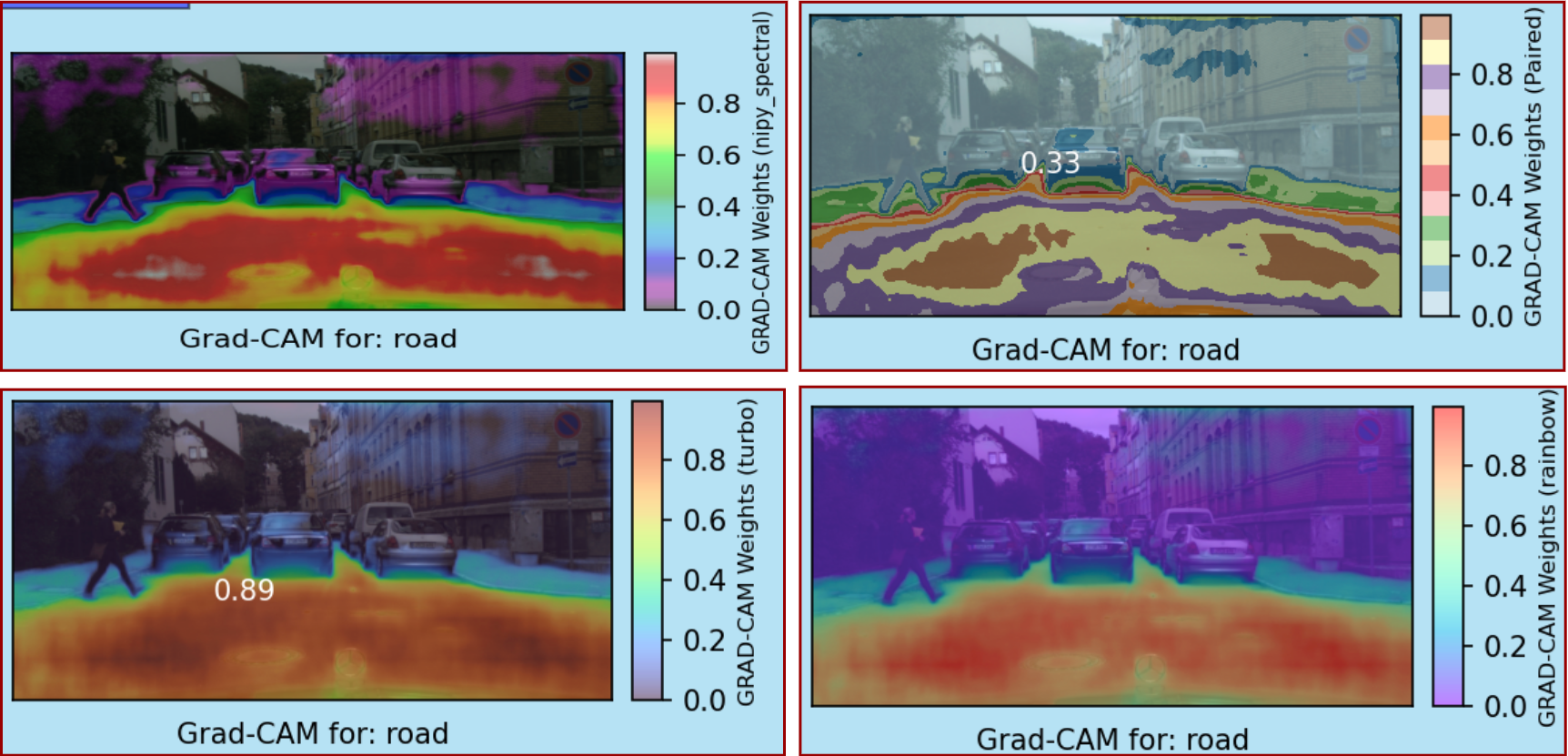}
    \caption{Cropped screenshot of our proposed tool for Task (section \ref{task_2}) from Jupyter notebook. Top Left: GRAD-CAM image with categorical (color-hue) colormap: nippy-spectral. Top Right: GRAD-CAM image with categorical (color-hue) colormap: Paired. Bottom Left: GRAD-CAM image with sequential colormap: Turbo. Bottom Right: GRAD-CAM image with sequential colormap: rainbow.}
    \label{fig:vis_2_ana}
\end{figure}

\begin{figure*}[t]
    \centering
    \includegraphics[width=\textwidth]{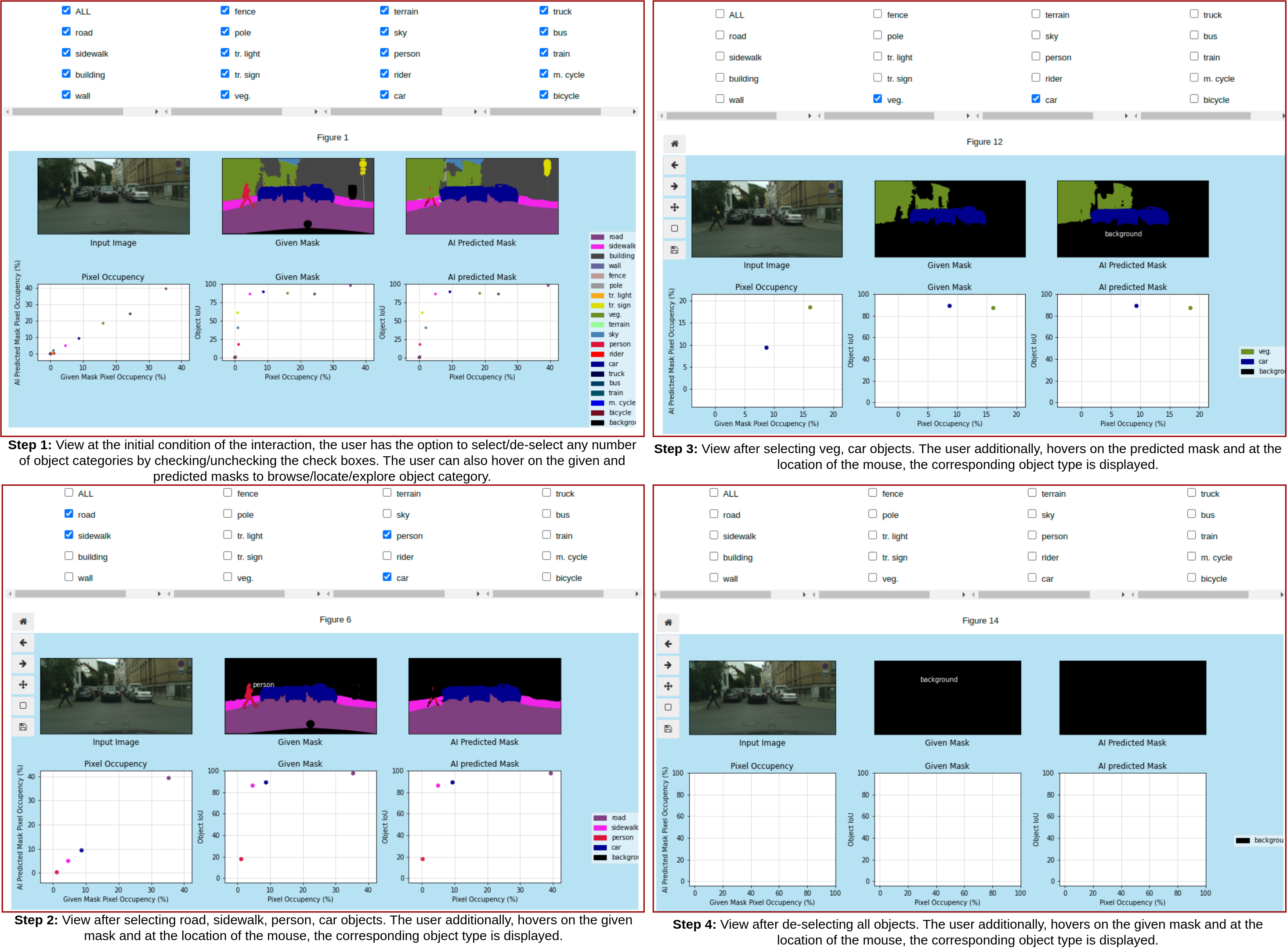}
    \caption{Cropped screenshot of our proposed tool for Task (section \ref{task_3}) from Jupyter notebook. Top Left: initial view of the proposed tool with no interaction. Bottom Left and Top Right: Original image, given mask, AI predicted mask, 3 scatter plots - (1) given and predicted pixel occupancy, (2) IoU and given pixel occupancy, and (3) IoU and predicted pixel occupancy for selected objects with hovers at the given mask (Bottom Left) and predicted mask (Top Right). Bottom Right: View after deselecting all objects with hover at the given mask.}
    \label{fig:vis_3}
\end{figure*}

Additionally, if the user attempts to locate a particular color (from the displayed color bar) in the image, they can do so by searching through the entire superimposed layer (Figure~\ref{fig:vis_2}). How easy or difficult it is for the user to do so is highly contingent on the choice of the colormap. For example, if the user uses sequential color maps such as 'paired', locating a color hue becomes much easier. In Figure~\ref{fig:vis_2_ana}, it can be observed that locating the second highest ranked color bin on the top right screen capture (categorical color map) is much easier than it is in the bottom left screen capture (sequential color map) where a sequential color map ('turbo') has been used. However, when the user focuses on exploring features which for this task translates to exploring which regions of the image are more important than the other when the AI model tries to find a certain object, sequential colormap ('turbo', 'viridis) are more user friendly (Figure~\ref{fig:vis_2_ana}). This advantage stems from the superimposed layer with sequential colormap yields a smoother distribution of color hue and, therefore, gradcam weights, as seen in Figure~\ref{fig:vis_2_ana}. Our recommended colormap for the user to explore features is turbo \cite{googleblogTurboImproved} as it is more perceptually uniform and therefore imposes a lesser cognitive load on the user.

Interestingly, while using our proposed tool for task 2, it can be further observed (Figure~\ref{fig:vis_2}) that when the AI model tries to predict bicycles, it looks at a portion of the image (legs of a person) that has some level of pseudo-similarity with the shape of bicycles (two rods connecting the wheels). This is a feature explored by decoding our design that may be of interest to the users (computer vision, AI researchers) while using this tool.

One limitation of our current design (Figure~\ref{fig:vis_2}, \ref{fig:vis_2_ana}) for this task is that AI attention weight can be explored/browsed/located for one object category at a time. Though it can be posited that viewing weights for too many objects at a time can be a high-load task, it may as well be desirable for particular users to explore/browse/locate grad-cam features for more than one category at a time. Moreover, running the gradcam algorithm \cite{selvaraju2017grad} is time-consuming, which is why we have restricted our tool (Figure~\ref{fig:vis_2}) to view grad-cam weights for one object category at a time since viewing multiple categories at once requires running the gradcam algorithm multiple times as well. 

Data transformation performed for this task (described in section \ref{task2_trans}) improves the match between data semantics and the task because exploring/browsing/locating grad-cam features would not have been possible without the transformation (running grad-cam algorithm on the original data and training AI on the original data). The interactions tool that we use here (Figure~\ref{fig:vis_2}, \ref{fig:vis_2_ana}) is effective because the user can locate grad-cam weights only by hovering. Also, while the user can locate the gradcam feature with the help of a color map and color bar, the hover gives the user the additional ability to locate grad-cam features more easily as the user does not have to depend on the color map while hovering (Figure~\ref{fig:vis_2_ana}). Moreover, we have used the horizontal spatial position channel, vertical spatial position channel to denote each image pixel, and the color-hue/color-saturation (depending on the chosen color map) channel is used to show (Figure~\ref{fig:vis_2}) the image and grad-cam weight (superimposed with image). Therefore, eliminating any of these channels won't allow us to show the aforementioned images and superimposed layers, which corroborates our hypothesis of "no redundant marks and channels".

\subsection{Task 3: Discovering Correlation Between Object Size and Object IoU for Different Category Classes} \label{task_3}

\subsubsection{Proposed Artifact}
While using this proposed tool (Figure~\ref{fig:vis_3}), users' goal is to locate, browse, and explore object categories in segmented images from a given image in the dataset. In our multiform view, we first display the input image, the corresponding given mask, and AI predicted mask. Additionally, this tool also provides (with three scatter plots) the user the ability to discover correlation among three pairs of attributes: (1) given and predicted pixel occupancy, (2) IoU and given pixel occupancy, and (3) IoU and predicted pixel occupancy for a given image, it's mask and AI predicted mask. We have used check boxes for all object categories in the cityscapes dataset. The user can select or deselect any of them at any given time. The information (Image, given mask, predicted mask, 3 scatter plots) is updated into the view (multiform) according to the objects selected.

\subsubsection{Data Abstraction}
The dataset type for this task is a spatial field (images and masks) and table (computed IoU, pixel occupancy for different objects in an image). While the data type for images and masks is grid, position, and attribute, the data types for IoU and pixel occupancy for different objects are items and attributes. In the table, each item is an object mask in the chosen image. And there are three attributes: (1) categorical attribute: object type, (2) ordinal attribute: object IoU, and (2) ordinal attribute: object size. We use a point mark and one horizontal position channel to encode pixel occupancy, a vertical position channel to encode object IoU in one view and pixel occupancy in another, and a color-hue channel to encode object category (class). For the images and masks, there are four channels: (1) horizontal spatial position, (2) vertical spatial position, (3) color saturation (while showing images), and (4) color hue (while showing masks). The mark used in this visualization is area.

\subsubsection{Task Abstraction}
When the user hovers over a given or predicted mask, the goal is to \textbf{locate}, \textbf{browse}, and \textbf{explore} \textbf{distribution}. In the case when they hover over a particular object and the object name is displayed, the user locates objects. Then, when the user hovers over a particular location of the image and the corresponding object name displays, the user browses objects. Finally, the user explores objects when they hover over different locations of the image, and our tool displays the corresponding object category. Furthermore, the user task is to discover the correlation among (1) pixel occupancy (in the given mask) and object IoU, (2) pixel occupancy (in the prediction mask) and object IoU, (3) pixel occupancy (in the given mask) and pixel occupancy (in the prediction mask).

\subsubsection{Encoding and Interaction Idiom}
We use a multiform view encoding. As shown in Figure~\ref{fig:vis_3}, on the top left of the multi-form view, we show an image; on the top middle, we show the given mask, and on the top right, we display the predicted mask and all of them are visually encoded in the form of an image. Then, we show three correlations on the bottom and use three scatter plots to encode them. Our design has multiple interaction choices. Firstly the user can select any number of object categories using the checkboxes that we implement. Then the user can also hover (select and highlight) the prediction and given masks. The initial overview shows the scatter plots, images, and masks for all the object categories in the cityscapes dataset.

\subsubsection{Critical Analysis}
One of the principal goals of this task is to locate/browse/explore object categories in the image by looking at given and predicted masks. While looking at the predicted and given a mask with our tool (Figure~\ref{fig:vis_3}), the users can discern different objects with the user of color hue, though this can be difficult sometimes. In cases with too many objects in an image or the objects are densely meshed, our designed color hue may not be enough to tell different objects apart. We have incorporated the hover (select) option to remedy that situation. The users can easily locate/explore different objects by hovering since our tool displays the object category at each mouse position, as seen in Figure~\ref{fig:vis_3}.

Another efficacious design choice that we use for this tool is the multiform view. Since we are showing (Figure~\ref{fig:vis_3}) different information/data (6 in this case) with different idioms (2 in this case), this choice helps take the cognition load off of the user. And when there is information being displayed for too many object categories (how many is too many can be different for every image since some images may contain too many object categories meshed with each other, some may contain the same amount of objects but spaced apart evenly), decoding the object categories from the mask can be challenging. To remedy that, we introduce (Figure~\ref{fig:vis_3}) the ability to interact as the users can choose for which object categories they want our tool to load information.

It is further observed in Figure~\ref{fig:vis_3} that for all objects, some of the scatter plots are difficult to read, especially when many points are positioned densely (the last two scatter plots in the Top Left of Figure~\ref{fig:vis_3}). On the other hand, Figure~\ref{fig:vis_3} shows that the user can understand the correlation between pixel occupancy and IoU for different objects by looking at the scatter plots. Since the scatter plots exhibit a positive correlation (Figure~\ref{fig:vis_3}), it helps the users answer questions such as: Does occupying more pixels lead to higher AI performance? The first scatter plot in Figure~\ref{fig:vis_3} also shows if, for different objects, the pixel occupancy in the given mask is equal or more or less in comparison to that of the predicted mask.

Section \ref{task3_trans} describes the required data transformations to change the data semantics for task 3. These transformations are exceedingly crucial because, without the trained AI model, we cannot display AI-predicted masks and cannot compute IoU and pixel occupancy in the predicted mask, which are all visualized attributes (Figure~\ref{fig:vis_3}) in the proposed idiom. 

\begin{figure*}[t]
    \centering
    \includegraphics[width=\textwidth]{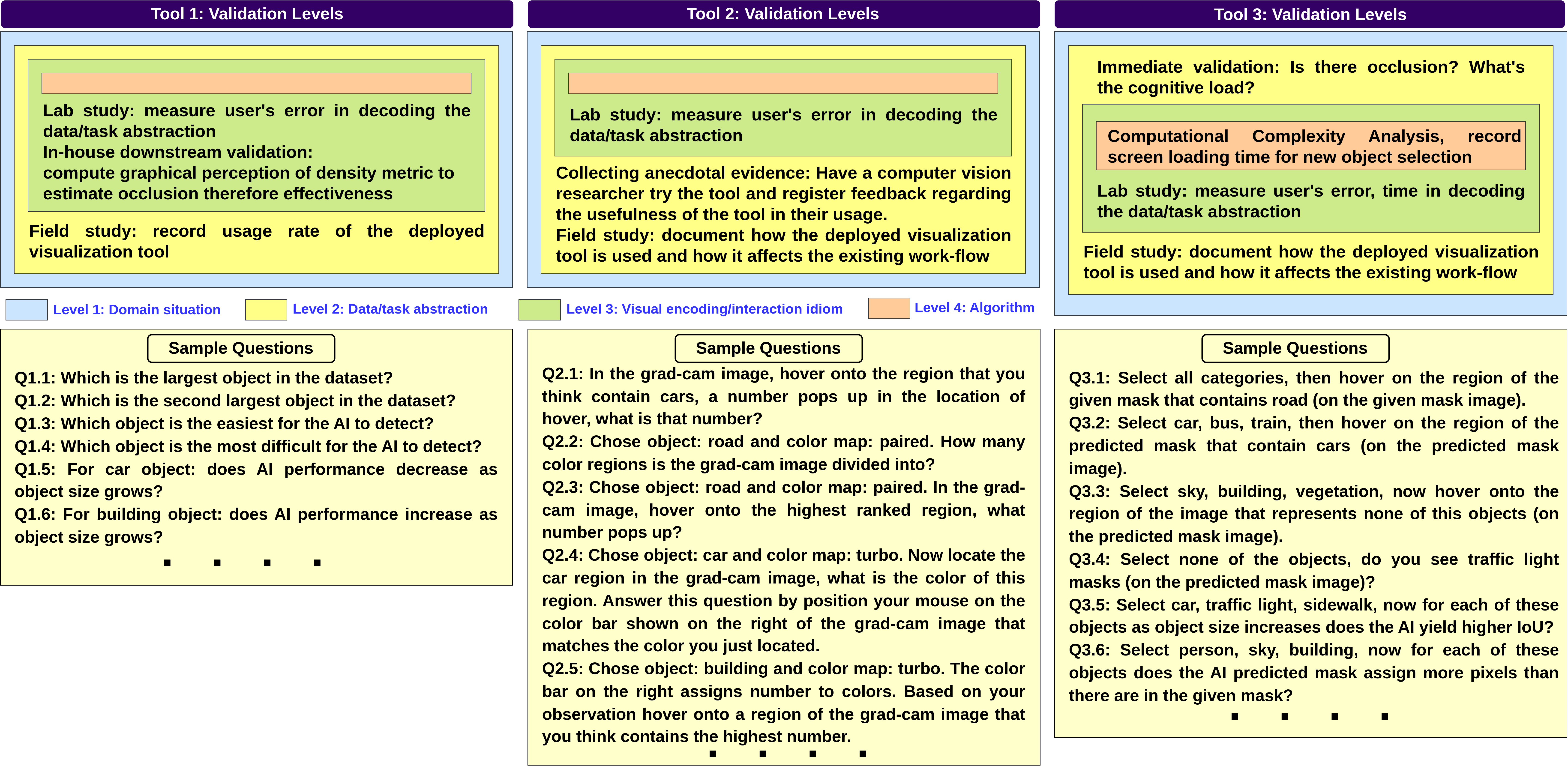}
    \caption{Four nested levels of visualization design with user validation sample questions for tools 1, 2, and 3 are shown on the left, middle, and right respectively. For each tool, the four nested levels with the required validation are on top, and on the bottom are the sample questions for those required validations.}
    \label{fig:valid}
\end{figure*}

\section{User Survey/Validation}
We further propose a set of validation mechanisms for our devised visualization tools with inspirations from \cite{robertson2008effectiveness, padilla2022multiple, nobre2020evaluating, franklin2017dashboard}. As Figure~\ref{fig:valid} depicts, we primarily focus on visual encoding/interaction idiom and data/task abstraction level for visualization tool 1 (described in Section~\ref{task_1}) and tool 2 (described in Section~\ref{task_2}) and further include algorithm level validation for tool 3 (described in Section~\ref{task_3}).

\subsection{Validating Task 1}
\subsubsection{Visual encoding/interaction idiom validation}
We propose a set of lab study questionnaire to validate the efficacy of our visualization tool for task 1 (described in Section~\ref{task_1}). Q3.1 to Q3.6 in Figure~\ref{fig:valid} depicts some of the sample (high-level) questions that determine the user's understanding of the correlation between IoU and object size. We aim to compute the average accuracy of these questions (Q3.1 to Q3.6 in Figure~\ref{fig:valid}) on a scale of 100 to measure the effectiveness of our tool in decoding information. We further propose to measure the average time (normalized to a scale of 100) taken by users to answer the questions (Q3.1 to Q3.6 in Figure~\ref{fig:valid}) in order to estimate our proposed tool's efficacy. Finally, we propose to take a weighted average of accuracy $a$ with the weight of $\alpha$ and measured time $t$ with the weight of $\beta$ to score ($s = (a \alpha + t \beta) / (\alpha + \beta) $) the effectiveness of our tool. The values of $\alpha$, and $\beta$ are subject to further experimentation. Additionally, as an in-house (downstream) idiom validation step, we further aim to quantitatively validate the designed tool, without the need of a user, by using the graphical perception of density metric \cite{dakin2011common} to quantify the occlusion/distinguish-ability between data points to discern patterns.

\subsubsection{Data / Task abstraction validation}
We propose a field study (Tool 1 in Figure~\ref{fig:valid}) to ask the target users to use our visualization tool in real-life situations, considering which visualization tool was developed. Example: (1) Registering if computer vision and machine learning researchers use this tool to gain insights at the end of each epoch/iteration of training. (2) Class imbalanced data requires training with many different models and mechanisms, increasing the time to develop a final model. We plan to observe if the usage of our tool can reduce the aforementioned time.

\subsection{Validating Task 2}
\subsubsection{Visual encoding/interaction idiom validation}  \label{valid_task2_scoring}
We propose a set of lab study questionnaires (Q2.1 to Q2.5 in Figure~\ref{fig:valid}) to examine how well users can decode information using our tool for task 2 (Section~\ref{task_2}). Q2.1 in Figure~\ref{fig:valid} determines if the user can browse the features of the grad-cam. On the other hand, Q2.3 in Figure~\ref{fig:valid} suggests if the user can locate grad-cam features, while Q2.5 in Figure~\ref{fig:valid} defines how well the user can browse grad-cam features. Along with the answers to these questions (correctness is defined by the answer's accuracy $a$), we further aim to record time $t$ (to study how fast the user can decode), number of mouse clicks $m$ (to understand it requires more interactions to answer certain questions than others which helps us understand how complex our visualization tool is) and user's confidence $c$ in answering each question. To record the user's confidence in each answer, we prompt five options: highly confident (value of $100$), confident (value of $80$), somewhat confident (value of $60$), not confident (value of $40$), not confident at all (value of $0$). We will then take a weighted average of these four quantities averaged over all questions and compute a score $s = (a \alpha + t \beta + (1-m) \gamma + c \delta) / (\alpha + \beta + \gamma + \delta)$ where $\alpha$, $\beta$, $\gamma$, $\delta$ are weights of $a$ (scale of 100), $t$ (normalized to a scale of 100), $(1-m)$ (normalized to a scale of 100), $c$ (scale of 100).

\subsubsection{Data / Task abstraction validation}
To collect anecdotal evidence (Tool 2 in Figure~\ref{fig:valid}), we plan to have a computer vision researcher try our tool and register qualitative feedback regarding the usefulness of the tool in their usage. Furthermore, we propose a field study (Tool 2 in Figure~\ref{fig:valid}) to ask the target users to use the visualization tool in their real-world workflow, observe and document how the deployed visualization tool is used, and register the changes (if any) occurring in their existing workflow. One way to do that is to observe if the users apply this tool more frequently when they train an AI model(s) for semantic segmentation.

\subsection{Validating Task 3}
\subsubsection{Algorithm validation}
We propose (Tool 3 in Figure~\ref{fig:valid}) to analyze  the computational complexity of the design using the following criterion. Every time the user selects a new set of object categories, new mask information is loaded on the screen that is specific to the selected objects. Currently (Figure~\ref{fig:vis_3}), we use \textit{for loop} to iteratively update the screen with user interaction. We plan to replace these iterative approaches with linear algebra operations on the segmentation mask tensors to examine if this is a bottleneck and if this proposed solution can reduce the computational load.

\subsubsection{Visual encoding/interaction idiom validation}
As a form of Immediate validation (Tool 3 in Figure~\ref{fig:valid}), we first investigate if the design occludes any information without involving the users for task 3 (Section~\ref{task_3}). And furthermore, we investigate if our design incorporating a multi-form view instead of switching between views has reduced cognitive load based on our judgment.

For the purpose of the lab study, we propose a set of questionnaires (Q3.1 to Q3.6 in Figure~\ref{fig:valid}). Q3.1 to Q3.4 in Figure~\ref{fig:valid} depicts the understanding of the user in terms of exploring/locating/browsing object categories in the segmented images (both by given mask and AI model). While Q3.5 to Q3.6 in Figure~\ref{fig:valid} determines the understanding of the user in discovering correlations between object distribution and IoU. We plan to use the scoring mechanism identical to Section~\ref{valid_task2_scoring} to determine the efficacy of our proposed visualization tool 3.

\subsubsection{Data / Task abstraction validation}
We propose a field study (Tool 3 in Figure~\ref{fig:valid}) to study our design's effectiveness. One of our proposed criteria to validate abstraction is to observe and record if the users are using this tool on research that is strictly related to training of AI models on class imbalanced semantic segmentation dataset; if so, the users are trying to compare the correlation between performance (IoU) and object frequency (pixel occupancy).

\section{Future Work}
One of the more pronounced stretch goals of our visualization tool is to facilitate the usage of our tool in the training phase. What currently prevents us from incorporating this feature is the computational complexity of our tool. To run the grad-cam algorithm during the training phase, which is computationally expensive, becomes an added stress to the user. Because, usually, while training AI models, most of the hardware memory is occupied by models and data loaders, allowing computational resources to the visualization is challenging. More sophisticated re-factorization of our code to ensure optimal hardware usage is required to solve the aforementioned issue. If Incorporated, this feature will help the user gain insights into the model's behavior as they train models. Furthermore, this feature will help the user understand how model weights change during training. This visualization can, therefore, potentially become a helping tool towards demystifying the infamous \textit{black-box} issues \cite{rudin2019stop} in machine learning. 

Another stretch goal of ours is applying our visualization tool to additional datasets and different types of AI models. We have only experimented on CNN-based AI models; the next candidate AI model we plan to experiment with is vision-transformer \cite{dosovitskiy2020image} based AI models. Then we plan to expand our experimentation on different semantic segmentation datasets from various application sub-domains such as autonomous vehicle application: Scannet \cite{dai2017scannet}, GTA5 \cite{richter2016playing}, IDD \cite{varma2019idd} and medical application: Kvasir-SEG \cite{jha2020kvasir}, PROMISE12 \cite{litjens2014evaluation}. Additional experiments such as those mentioned above can potentially yield more insights into our visualization's usability, efficacy, and adaptability in different domains.

In terms of usage, one of the biggest limitations of our visualization tool is its adaptability to another dataset by a different user. Developing a web portal or a generic Python library can potentially solve that problem. A user can easily use the web portal without configuring any new environment on their system and can use our visualization from the server. On the other hand, developing a Python library will help users integrate the visualization into their project easily with some basic configuration changes due to their (the user's) dataset and AI model. To mitigate these challenges, we require proper documentation of our code, which can further attract more contributors to our GitHub repository to drive project growth, quality, and usability.

Another limitation of our visualization tool is that users can't select or upload images for the tasks described in Section~\ref{task_2}, \ref{task_3}. Giving that level of control to users will enable them to use our visualization tool for demonstration purposes more easily; our tool can be viewed as a readily available platform for the aforementioned tasks. Finally, there is always more room for making the visualizations more interactive, and these decisions and design policies need to be made with extensive research, prototyping, and the adoption of user feedback.

\section{Conclusion}
We have presented three visualizations to understand AI interpretability and explore semantics between dataset statistics and AI models in a novel way. We further propose a user survey that can potentially help us understand how our proposed visualization can help target users better correlate their data and models. With learning incorporated effectively from user feedback and surveys, we hope that our visualization tool will be an efficacious choice in completing the aforementioned tasks driven by our visualization tool in our target user's real-life workflow. We believe that with the incorporation of our stretch goals, our proposed visualization tool can potentially be a significant research tool in computer vision.

\section{Acknowledgement}
The authors would like to express gratitude to Professor Rebecca Williams for their guidance during the course that led to this paper's development. Their expertise and support greatly contributed to the quality of this research.

\bibliographystyle{IEEEtran}
\bibliography{main}

\end{document}